\documentclass{article}
\usepackage{spconf,amsmath,epsfig}
\usepackage{amsmath}
\usepackage{amssymb}
\usepackage{upgreek}
\usepackage{amsfonts}
\usepackage{makecell}
\usepackage{array}
\usepackage{units}
\usepackage{multirow}
\usepackage{booktabs}
\usepackage{makecell}
\usepackage{tabularx}
\usepackage{url}
\usepackage[square,numbers]{natbib}

\title{A Novel Graph-theoretic deep representation learning Method for multi-label remote sensing image retrieval}

\name{Gencer Sumbul and Beg\"{u}m Demir}
\address{Faculty of Electrical Engineering and Computer Science, Technische Universit\"at Berlin, Germany}

\begin{document}
%
\maketitle
\begin{abstract}
This paper presents a novel graph-theoretic deep representation learning method in the framework of multi-label remote sensing (RS) image retrieval problems. The proposed method aims to extract and exploit multi-label co-occurrence relationships associated to each RS image in the archive. To this end, each training image is initially represented with a graph structure that provides region-based image representation combining both local information and the related spatial organization. Unlike the other graph-based methods, the proposed method contains a novel learning strategy to train a deep neural network for automatically predicting a graph structure of each RS image in the archive. This strategy employs a region representation learning loss function to characterize the image content based on its multi-label co-occurrence relationship. Experimental results show the effectiveness of the proposed method for retrieval problems in RS compared to state-of-the-art deep representation learning methods. The code of the proposed method is publicly available at \url{https://git.tu-berlin.de/rsim/GT-DRL-CBIR}.
\end{abstract}
\vspace{-0.03in}
\section{Introduction}
\vspace{-0.03in} 
Multi-label content-based image retrieval (CBIR) methods aim to retrieve remote sensing (RS) images similar to a given query image by exploiting training images annotated by multi-labels. Development of effective CBIR methods has recently attracted great attention in RS. As an example, in \citep{Dai:2018} a sparse reconstruction-based multi-label RS image retrieval method that considers a measure of label likelihood is introduced. In \cite{Shao:2020}, fully convolutional networks are introduced for multi-label RS images to extract descriptors of image regions in the content of CBIR. Recently, deep representation learning (DRL) methods based on a triplet loss function are found very popular for CBIR problems due to their intrinsic characteristic to model similarities of images. These methods employ image triplets (each of which includes anchor, positive and negative images), aiming to learn a metric space where the distance between the positive and the anchor images is minimized while that between the negative and anchor images is maximized. In \cite{Roy:2020}, triplet loss is employed with convolutional neural networks (CNN) to learn an embedding space for hash code generation of RS images. The use of triplet loss function requires an accurate selection of image triplets. A simple strategy is to define triplets from an existing training set of labeled images. However, such strategy does not guarantee the selection of the most informative images to the anchor, and thus can result in limited CBIR performance particularly when images annotated by multi-labels are available. In addition, the triplet selection based DLR methods do not take into account the co-occurrence relationships of land-cover classes present in an RS image. However, modeling these relationships is crucial for an accurate CBIR. This problem can be addressed by using graphs, which capture both region characteristics and the spatial relationships among the regions. In \citep{Chaudhuri:2018}, a semi-supervised graph-theoretic method is introduced to model inherent correlation of multi-labels by a correlated label propagation algorithm. The performance of this approach depends on the hand-crafted features to represent each image region. Recently, in \cite{Chaudhuri:2019} region graph-based image representations are utilized to model the similarity of image pairs via a siamese graph CNN in the context of DRL. This method learns a metric space based on only pairwise image similarities, which may not be sufficient to model the complex information content of RS images for CBIR problems. 

To address the above-mentioned issues, in this paper we propose a graph-theoretic deep representation learning method that does not require image pairs and triplets. The proposed method models multi-label co-occurrence relationships based on a novel region representation learning loss function.
\vspace{-0.03in}
\section{The Proposed Graph-Theoretic Deep Representation Learning Method}
\vspace{-0.03in}
\begin{figure*}[t]
  \centering
  \includegraphics[width=0.97\linewidth]{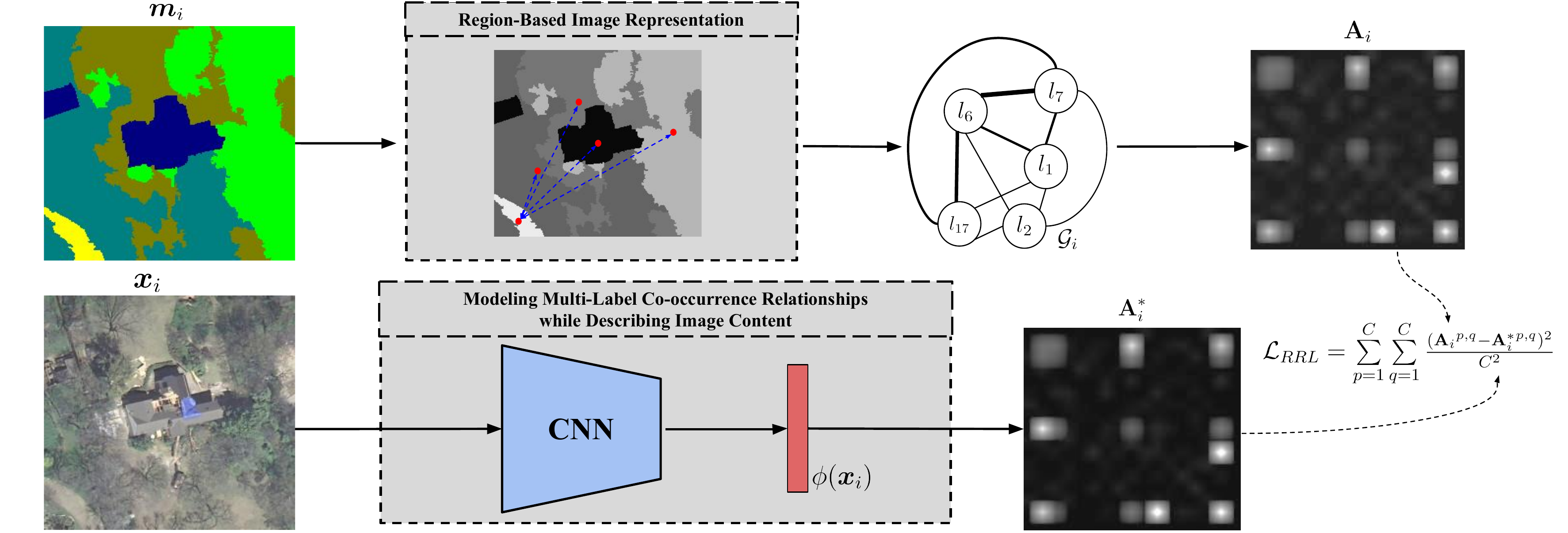}
  \caption{Illustration of the proposed graph-theoretic deep representation learning method.}
  \label{fig:model}
  \vspace{-0.3cm}
\end{figure*}
Let \mbox{$\mathcal{X}{=}\{\boldsymbol{x}_1,\ldots,\boldsymbol{x}_I\}$} be an archive that includes $I$ images, where $\boldsymbol{x}_j$ is the $j$\textsuperscript{th} RS image in the archive $\mathcal{X}$. We assume that a training set $\mathcal{T} \subset \mathcal{X}$ that consists of labeled images is available. Each image in $\mathcal{T}$ is associated to pixel-based labels from a label set \mbox{$\mathcal{B} = \{l_1,...,l_C\}$}. Let $\boldsymbol{m}_i$ be the land-cover map of the image $\boldsymbol{x}_i \in \mathcal{T}$ ($\boldsymbol{m}_i$ and $\boldsymbol{x}_i$ have the same pixel sizes and thus each pixel in $\boldsymbol{m}_i$ represents the label of the corresponding pixel in $\boldsymbol{x}_i$). The set of all labels associated to $\boldsymbol{x}_i$ are defined by a binary vector $\boldsymbol{y}_i \in \{0,1\}^{C}$, where each element of $\boldsymbol{y}_i$ indicates the presence or absence of label $l_n \in \mathcal{B}$. 

The proposed method aims to model co-occurrence relationship of multiple classes present in each image in the archive. To this end, each training image $\boldsymbol{x}_i$ is represented with a graph structure, which provides region-based image representation (where each region is associated with a land-cover class). The proposed method includes a novel learning strategy to automatically predict the corresponding graph structure of any image in the archive, while describing the complex content of each image. To this end, we exploit a convolutional neural network (CNN). However, the proposed learning strategy can be injected to any deep neural network. Fig. \ref{fig:model} shows a general overview of the proposed method, which is explained in detail in the following.

During training, to describe regions associated to classes present in each training image, the proposed method first constructs a graph structure, where the nodes represent the image region properties and the edges represent the spatial relationship among the regions. Let $\mathcal{G}_i = (E_i, V_i, \mathbf{W}_i)$ be the graph associated to the image $\boldsymbol{x}_i$. $E_i$ is the set of graph edges, $V_i$ is the set of nodes and $\mathbf{W}_i \in \mathbb{R}^{C \times C}$ is the weight matrix of the graph. Each node represents a region associated to a class of the image (i.e., $l_n \in \boldsymbol{y}_i$). The weight $\mathbf{W}_i^{p,q}$ between $l_p$ and $l_q$ $\mathbf{W}_i^{p,q}=1$ if $l_p \in \boldsymbol{y}_i, l_q \in \boldsymbol{y}_i$, and otherwise $\mathbf{W}_i^{p,q}=0$.

By this way, all the class relationships of $\boldsymbol{x}_i$ are modeled with same importance. However, class relationships can be subject to different levels of importance based on the characteristics of each region and its spatial relationship with the other regions. As an example, the relationships between an image region and its neighbors are more important than those between non-neighbor regions. In detail, if two neighbor regions cover most of the image content, their relationship plays the most significant role for accurately modeling the multi-label co-occurrence relationship. To address this issue, we define a weight for the edge $\mathbf{W}_i^{p,q}$ as follows:
\begin{equation} \label{eq1}
\mathbf{W}_i^{p,q} = \frac{s(l_p; \boldsymbol{m}_i) \times s(l_q; \boldsymbol{m}_i)}{N_s} \times (1 - \frac{d(l_p, l_q; \boldsymbol{m}_i)}{N_d})
\end{equation}
where $s: \mathcal{B} \mapsto \mathbb{N}$ is a function that maps a class label into the size of the region associated to the class, $d: \mathcal{B} \times \mathcal{B} \mapsto \mathbb{N}$ is a function that maps the pairs of class labels into the distance between the centers of their regions associated to the corresponding classes. $N_s$ and $N_d$ are the maximum values of the functions $s$ and $d$, respectively. By this way, if the regions are close to each other and their sizes are large, the weights assigned to the corresponding edges in the graph $\mathcal{G}_i$ will be high. After obtaining a graph for each training image, the characteristics and spatial arrangements of image regions are represented with an adjacency matrix $\mathbf{A}_i \in \mathbb{R}^{C \times C}$ where $\mathbf{A}_i^{p,q}=\mathbf{W}_i^{p,q}$ if an edge exists between the nodes $V_i^p$ and $V_i^q$ in the graph $\mathcal{G}_i$, $\mathbf{A}_i^{p,q}=0$ otherwise.

To model multi-label co-occurrence relationship of any image in the archive, the proposed learning strategy consists of region-based image representation learning and image characterization. Let $\phi : \theta, \mathcal{X} \mapsto \mathbb{R}^{\gamma}$ be any type of CNN that maps the image $\boldsymbol{x}_i$ to $\gamma$-dimensional image descriptor, where $\theta$ is the set of CNN parameters. The region-based image representation learning is achieved by the prediction of the adjacency matrix based on the image descriptor. To this end, the characterization of $\boldsymbol{x}_i$ is performed based on the considered CNN to model the multi-label co-occurrence relationship of $\boldsymbol{x}_i$ in the adjacency matrix $\mathbf{A}_i$. The prediction of the adjacency matrix is achieved by a fully connected layer that takes the image descriptor $\phi(\boldsymbol{x}_i)$ and produces the vectorized form of the reconstructed adjacency matrix. To train the proposed method, we define a novel region representation learning loss $\mathcal{L_{RRL}}$ function as follows:
\begin{equation}
 \setlength\abovedisplayskip{3pt}
    \mathcal{L_{RRL}} = \sum_{\boldsymbol{x}_i\in \mathcal{T}}\sum_{p=1}^C\sum_{1=1}^C \frac{({\mathbf{A}_i^{p,q} - \mathbf{A}_i^*}^{p,q})^2}{C^2}.
\end{equation}
The proposed loss function allows to describe the content of an RS image based on the multi-label co-occurrence information to achieve the region-based image representation learning. After an end-to-end training of the whole neural network by minimizing the region representation learning loss and thus learning the network parameters $\theta^* = \operatorname*{arg\,min}_\theta \mathcal{L_{RRL}}$, the proposed method extracts the descriptors $\{\phi(\boldsymbol{x}_j;\theta^*)\}$ of the images in the archive $\mathcal{X}$. To perform CBIR, the proposed method retrieves RS images from the archive similar to a given query image $\boldsymbol{x}_q$ by comparing $\phi(\boldsymbol{x}_q)$ with each element of the set $\{\phi(\boldsymbol{x}_j;\theta^*) \}$. 

It is worth noting that the proposed method considers $\boldsymbol{m}_i$ of $\boldsymbol{x}_i \in \mathcal{T}$ (i.e., pixel-level labels of training images) is available for the training phase. In the case that training images are annotated by image-level multi-labels instead of pixel-level labels, $\boldsymbol{m}_i$ can be obtained by using a weakly-supervised semantic segmentation that exploits only image-level annotations as explained in \cite{Chan:2020}.
\vspace{-0.03in}
\section{Experimental Results}
\vspace{-0.03in}
Experiments were conducted on the DLRSD \cite{DLRSD} and the BigEarthNet-S2 \cite{BigEarthNet} benchmark archives. The DLRSD archive is the extension of the UC Merced archive \cite{Yang:2013} that includes 2,100 aerial images, each of which has the size of 256 $\times$ 256 pixels with a spatial resolution of 30 cm. The DLRSD archive also includes pixel labels defined in \cite{Chaudhuri:2018}. To perform experiments, we split the DLRSD archive into training (80\%) and test (20\%) sets. The large-scale BigEarthNet-S2 benchmark archive consists of 590,326 Sentinel-2 images. Each image in BigEarthNet-S2 has been annotated with multi-labels from the 2018 CORINE Land Cover (CLC) database. In this paper, we first extracted the CLC land cover map of each image and then exploited it based on the 19 classes nomenclature presented in~\cite{BigEarthNet19}. 
\begin{table}[t] 
\renewcommand{\arraystretch}{0.1}
\centering
\caption{Mean average precision (mAP) obtained for the DLRSD and BigEarthNet-S2 archives.}
\label{table:map}
\begin{tabular}{
@{\hskip 0.025in}*{1}{>{\raggedright\arraybackslash}p{0.17\textwidth}}*{1}{>{\centering\arraybackslash}p{0.12\textwidth}}*{1}{>{\centering\arraybackslash}p{0.128\textwidth}}}
\toprule
\midrule
\multirow{2}{0.23\textwidth}[2pt]{\raggedright Method} & \multicolumn{2}{c}{Benchmark Archive} \tabularnewline \cmidrule{2-3}
 & DLRSD & BigEarthNet-S2 \tabularnewline
\toprule
SNN (random)~\cite{Schroff:2015} & 66.5\% & 83.9\% \tabularnewline \cmidrule{1-3}
SNN (batch-all)~\cite{Schroff:2015} & 68.0\% & 88.6\% \tabularnewline \cmidrule{1-3}
SNN (hard)~\cite{Schroff:2015} & 70.7\% & 88.3\% \tabularnewline \cmidrule{1-3}
SGCN~\cite{Chaudhuri:2019} & 70.1\% & 87.8\% \tabularnewline \cmidrule{1-3}
Proposed Method & \textbf{84.3\%} & \textbf{92.1\%} \tabularnewline \midrule
\bottomrule
\end{tabular}
\vspace{-0.4cm}
\end{table}
To perform experiments, we first selected the 74,716 BigEarthNet-S2 images acquired over Serbia and then divided them into training (52\%),
validation (24\%) and test (24\%) sets. 
To select query images, the training set of the DLRSD archive and the validation set of the BigEarthNet-S2 archive were used, while images were retrieved from the test set for both archives. In the experiments, we exploited the DenseNet model \cite{Huang:2017} at the depth of 121. We trained our method for 100 epochs by using the Adam optimizer. We compared our method with siamese neural networks (SNNs) trained with triplet loss~\cite{Schroff:2015} and siamese graph convolution network (SGCN) trained with contrastive loss~\cite{Chaudhuri:2019}. For SNN, we utilized random, batch-all and hard sampling techniques in the experiments. The results are denoted as SNN (random), SNN (batch-all) and SNN (hard). The reader is referred to \cite{Hermans:2017} for the details of these techniques. The same training procedure and the same backbone with the proposed method were used for all experiments. For SGCN, we employed the same graph formation and parameter values given in~\cite{Chaudhuri:2019}. To obtain CBIR results, chi-square distance is utilized to compare image descriptors. Experimental results are provided in terms of normalized discounted cumulative gains (NDCG), mean average precision (mAP) and average cumulative gains (ACG) \cite{Zhang:2020}. Table \ref{table:map} shows the mAP results obtained on both archives. By assessing the table, one can observe that the proposed method leads to the highest mAP scores compared to the SNN with all types of sampling techniques and SGCN. As an example, the proposed method provides almost 18\% higher and more than 8\% higher mAP scores for DLRSD and BigEarthNet-S2 archives, respectively, compared to the SNN (random). This shows that modeling multi-label co-occurrence of an RS image by the proposed method improves the CBIR performance compared to the SNN, in which multi-label dependencies present in an image have not been considered. As an other example, the proposed method provides almost 14\% higher mAP score for the DLRSD archive compared to the SNN (hard). In addition, the proposed method leads to more than 14\% higher and almost 5\% higher mAP scores for DLRSD and BigEarthNet-S2 archives, respectively, compared to the SGCN, which is one of the state-of-the-art graph-based DRL methods for CBIR. These results show that, without a need for pair or triplet selection, the proposed method characterizes the image similarity much more accurately compared to the triplet and contrastive loss based DRL methods. 
\begin{figure}[!htb]
  \centering
 \begin{minipage}[t]{.5\textwidth}
  \centering
  \centerline{\includegraphics[width=0.97\textwidth]{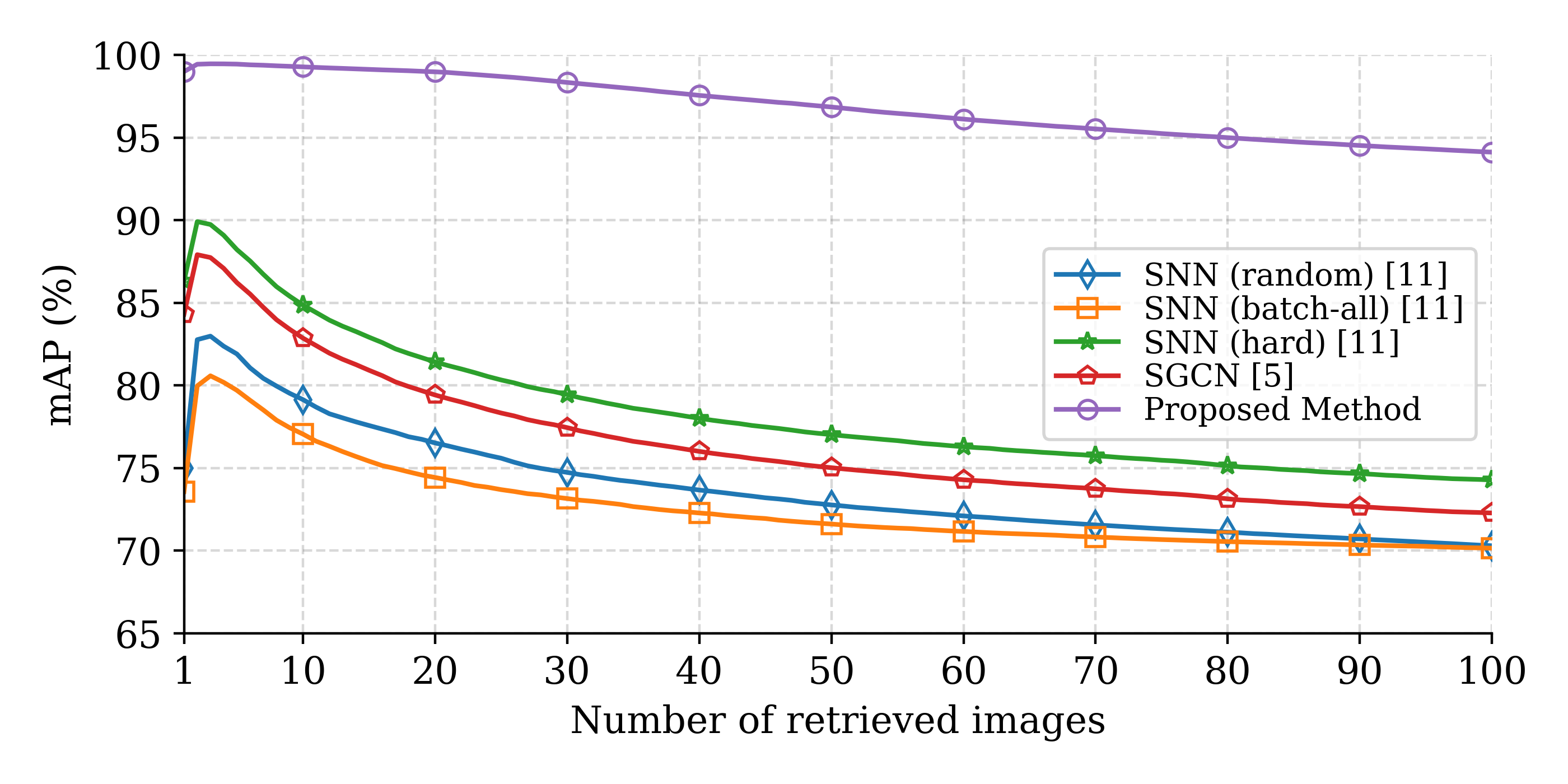}}
  \vspace{-0.36cm}
  \centerline{(a)}
  \vspace{-0.05cm}
\end{minipage}
\begin{minipage}[t]{.5\textwidth}
  \centering
  \centerline{\includegraphics[width=0.97\textwidth]{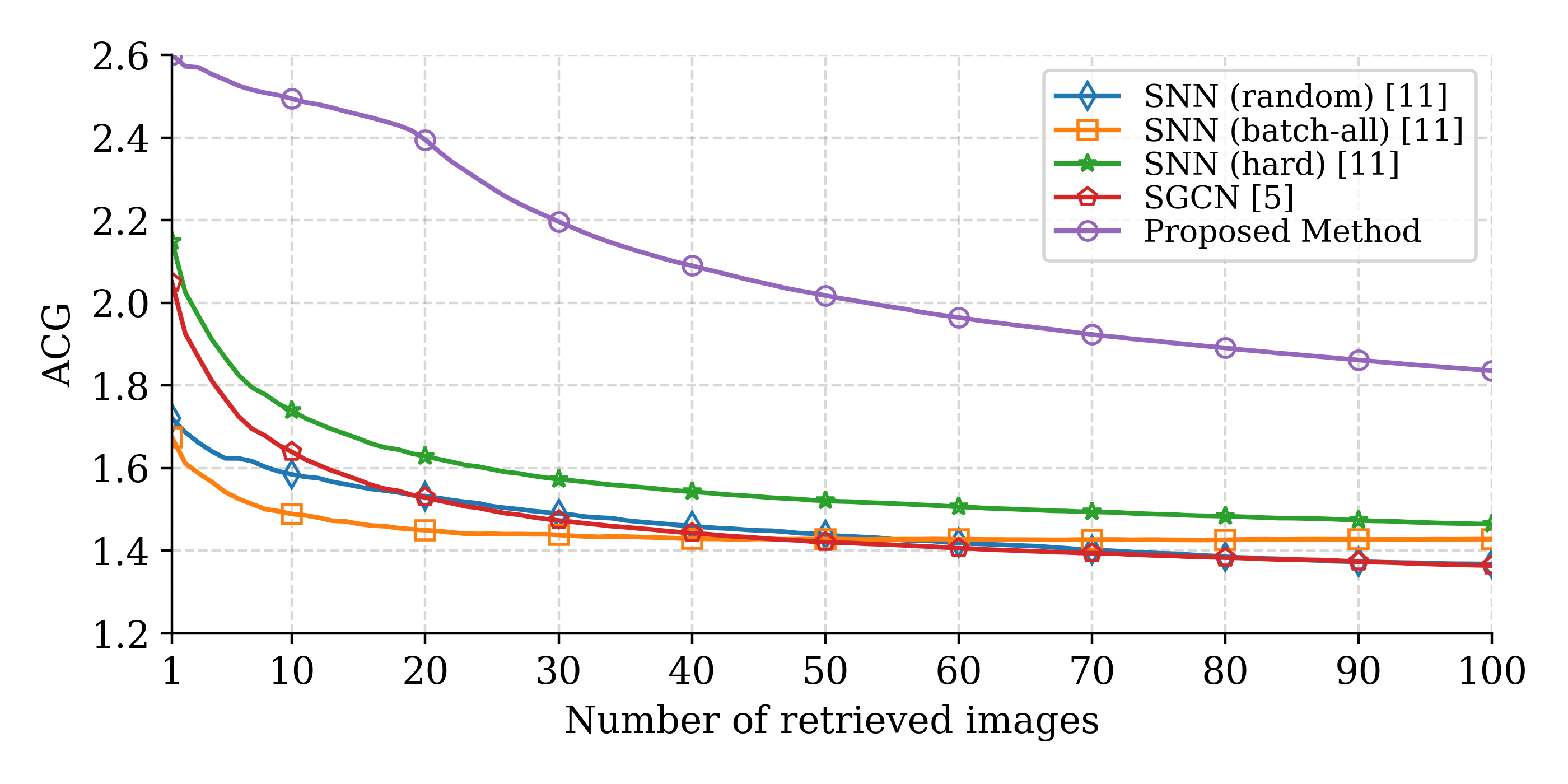}}
  \vspace{-0.36cm}
  \centerline{(b)}
  \vspace{-0.05cm}
\end{minipage}
\begin{minipage}[t]{0.5\textwidth}
  \centering
 \centerline{\includegraphics[width=0.97\textwidth]{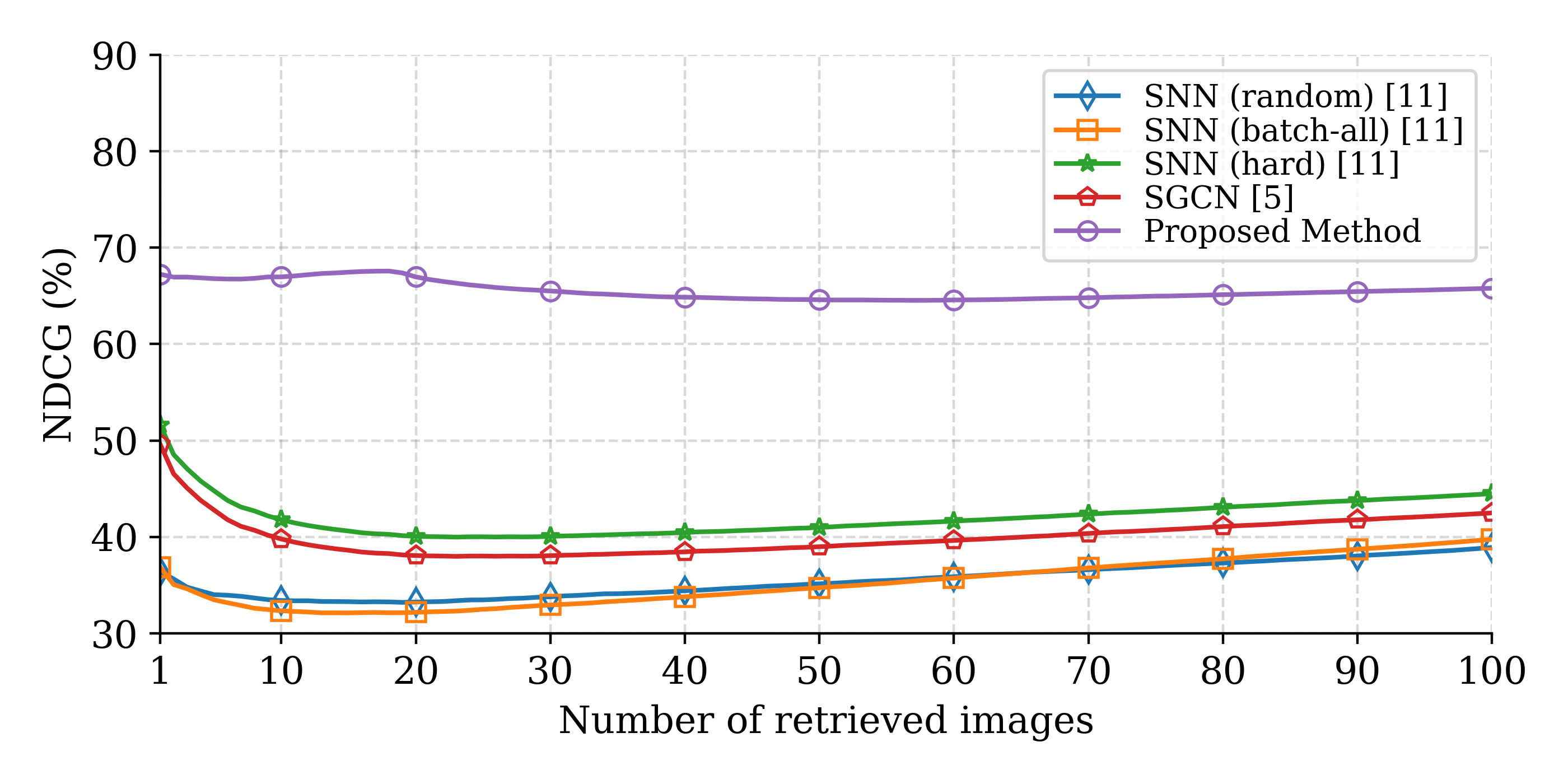}}
  \vspace{-0.36cm}
  \centerline{(c)}
  \vspace{-0.05cm}
\end{minipage}
\caption{(a) Mean average precision (mAP); (b) Average cumulative gains (ACG) and (c) Normalized discounted cumulative gains (NDCG) versus the number of retrieved images obtained for the DLRSD archive.}
\label{fig:res}
\vspace{-0.4cm}
\end{figure}
Fig. \ref{fig:res} shows the mAP, ACG and NDCG results for the DLRSD archive under different numbers of retrieved images. By analyzing the figure, one can see that increasing the number of retrieved images does not change our conclusion. As an example, the proposed method outperforms SCNN and SGCN by almost 20\% in NDCG for the DLRSD archive when the number of retrieved images is 100. It is worth noting that the promising CBIR performance of our method relies on: i) accurately predicting the graph structures of images in the archive; and ii) defining image descriptors based on the co-occurrence relationship of land-cover classes present in each image. 
\vspace{-0.04in}
\section{Conclusion}
\vspace{-0.03in}
In this paper, we have presented a novel graph-theoretic deep representation learning method for multi-label RS image retrieval problems. The effectiveness of our method relies on the efficient use of a novel learning strategy that contains a region representation learning loss function (which allows to model an RS image content based on the multi-label co-occurrence relationships). Experimental results show the success of the proposed method compared to state-of-art deep representation learning methods.

It is worth noting that the proposed method requires training images annotated by pixel-level labels. Such class labels can be attained through publicly available thematic products. However, class labels available through the thematic products can be noisy (incomplete, outdated, etc.), and thus their direct use may result in an uncertainty in the DL models and thus uncertainty in the CBIR performance. As a future work, we plan to develop label-noise robust graph-theoretic deep representation learning methods.
\vspace{-0.03in}
\section{Acknowledgements}
\vspace{-0.03in}
This work was supported by the European Research Council under the ERC Starting Grant BigEarth-759764.
\bibliographystyle{IEEEbib}
\small
\bibliography{defs,refs}

\begin{thebibliography}{10}

\bibitem{Dai:2018}
O.~E. Dai, B.~Demir, B.~Sankur, and L.~Bruzzone,
\newblock ``A novel system for content-based retrieval of single and
  multi-label high-dimensional remote sensing images,''
\newblock {\em IEEE J. Sel. Top. Appl. Earth Obs. Remote Sens.}, vol. 11, no.
  7, pp. 2473--2490, 2018.

\bibitem{Shao:2020}
Z.~{Shao}, W.~{Zhou}, X.~{Deng}, M.~{Zhang}, and Q.~{Cheng},
\newblock ``Multilabel remote sensing image retrieval based on fully
  convolutional network,''
\newblock {\em IEEE J. Sel. Top. Appl. Earth Obs. Remote Sens.}, vol. 13, pp.
  318--328, 2020.

\bibitem{Roy:2020}
S.~{Roy}, E.~{Sangineto}, B.~{Demir}, and N.~{Sebe},
\newblock ``Metric-learning-based deep hashing network for content-based
  retrieval of remote sensing images,''
\newblock {\em IEEE Geosci. Remote Sens. Lett.}, vol. 18, no. 2, pp. 226--230,
  2021.

\bibitem{Chaudhuri:2018}
B.~{Chaudhuri}, B.~{Demir}, S.~{Chaudhuri}, and L.~{Bruzzone},
\newblock ``Multilabel remote sensing image retrieval using a semisupervised
  graph-theoretic method,''
\newblock {\em IEEE Trans. Geosci. Remote Sens.}, vol. 56, no. 2, pp.
  1144--1158, 2018.

\bibitem{Chaudhuri:2019}
U.~Chaudhuri, B.~Banerjee, and A.~Bhattacharya,
\newblock ``Siamese graph convolutional network for content based remote
  sensing image retrieval,''
\newblock {\em Comput. Vis. Image Understand.}, vol. 184, pp. 22--30, 2019.

\bibitem{Chan:2020}
L.~Chan, M.~S. Hosseini, and K.~N. Plataniotis,
\newblock ``A comprehensive analysis of weakly-supervised semantic segmentation
  in different image domains,''
\newblock {\em Int. J. Comput. Vis.}, 2020.

\bibitem{DLRSD}
Z.~Shao, K.~Yang, and W.~Zhou,
\newblock ``Performance evaluation of single-label and multi-label remote
  sensing image retrieval using a dense labeling dataset,''
\newblock {\em Remote Sens.}, vol. 10, no. 6:964, 2018.

\bibitem{BigEarthNet}
G.~Sumbul, M.~Charfuelan, B.~Demir, and V.~Markl,
\newblock ``{BigEarthNet}: A large-scale benchmark archive for remote sensing
  image understanding,''
\newblock {\em IEEE Intl. Geosci. Remote Sens. Symp.}, pp. 5901--5904, 2019.

\bibitem{Yang:2013}
Y.~{Yang} and S.~{Newsam},
\newblock ``Geographic image retrieval using local invariant features,''
\newblock {\em IEEE Trans. Geosci. Remote Sens.}, vol. 51, no. 2, pp. 818--832,
  2013.

\bibitem{BigEarthNet19}
G.~Sumbul, A.~d.~Wall, T.~Kreuziger, F.~Marcelino, H.~Costa, P.~Benevides,
  M.~Caetano, B.~Demir, and V.~Markl,
\newblock ``{BigEarthNet-MM}: A large scale multi-modal multi-label benchmark
  archive for remote sensing image classification and retrieval,''
\newblock {\em arXiv preprint arXiv:2105.07921}, 2021.

\bibitem{Schroff:2015}
Florian S., Dmitry K., and James P.,
\newblock ``Facenet: A unified embedding for face recognition and clustering,''
\newblock in {\em IEEE Conf. Comput. Vis. Pattern Recog.}, 2015, pp. 815--823.

\bibitem{Huang:2017}
G.~{Huang}, Z.~{Liu}, L.~v.~d. {Maaten}, and K.~Q. {Weinberger},
\newblock ``Densely connected convolutional networks,''
\newblock in {\em IEEE Conf. Comput. Vis. Pattern Recog.}, 2017, pp.
  2261--2269.

\bibitem{Hermans:2017}
A.~Hermans, L.~Beyer, and B.~Leibe,
\newblock ``{In Defense of the Triplet Loss for Person Re-Identification},''
\newblock {\em arXiv preprint arXiv:1703.07737}, 2017.

\bibitem{Zhang:2020}
Z.~{Zhang}, Q.~{Zou}, Y.~{Lin}, L.~{Chen}, and S.~{Wang},
\newblock ``Improved deep hashing with soft pairwise similarity for multi-label
  image retrieval,''
\newblock {\em IEEE Trans. Multimedia}, vol. 22, no. 2, pp. 540--553, 2020.

\end{thebibliography}

\end{document}